\documentclass[conference]{IEEEtran}
\IEEEoverridecommandlockouts
\usepackage{cite}
\usepackage{amsmath,amssymb,amsfonts}
\usepackage{algorithmic}
\usepackage{graphicx}
\usepackage{textcomp}
\usepackage{xcolor}
\usepackage[caption=false]{subfig}
\usepackage{booktabs}
\usepackage{bm}
\newtheorem{property}{Property}

\def\BibTeX{{\rm B\kern-.05em{\sc i\kern-.025em b}\kern-.08em
		T\kern-.1667em\lower.7ex\hbox{E}\kern-.125emX}}
\begin{document}
	
	\title{ Hybrid Visual Servoing Tracking Control of Uncalibrated Robotic Systems for Dynamic Dwarf Culture Orchards Harvest\\	
		\thanks{
			This work was funded by Beijing Science and Technology Plan Project (No.Z201100008020009), China Postdoctoral Science Foundation(No.2020M680445), Postdoctoral Science Foundation of Beijing Academy of Agriculture and Forestry Sciences of China (No.2020-ZZ-001).}
	}

	\author{\IEEEauthorblockN{1\textsuperscript{st} Tao Li}
		\IEEEauthorblockA{\textit{Beijing Research Center of } \\
			\textit{Intelligent Equipment}\\
			\textit{ for Agriculture}\\
			\textit{Beijing Academy of Agriculture}\\
			\textit{and Forestry Sciences}\\
			Beijing, China \\
			lit@nercita.org.cn\\
			ORCID:0000-0002-7522-5090}
		\and
		\IEEEauthorblockN{2\textsuperscript{nd} Quan Qiu}
		\IEEEauthorblockA{\textit{Beijing Research Center of } \\
			\textit{Intelligent Equipment}\\
			\textit{ for Agriculture}\\
			\textit{Beijing Academy of Agriculture}\\
			\textit{and Forestry Sciences}\\
			Beijing, China \\		
			qiuq@nercita.org.cn\\
			ORCID:0000-0002-7261-3856}
		\and
		\IEEEauthorblockN{3\textsuperscript{rd} Chunjiang Zhao}
		\IEEEauthorblockA{\textit{National Engineering Research Center} \\
			\textit{ for Information Technology}\\
			\textit{ in Agriculture}\\
			\textit{Beijing Academy of Agriculture}\\
			\textit{and Forestry Sciences}\\
			Beijing, China \\
			zhaocj@nercita.org.cn\\
		}
		
	}
	
	\maketitle
	
	\begin{abstract}
		The paper is concerned with the dynamic tracking problem of SNAP orchards harvesting robots in the presence of multiple uncalibrated model parameters in the application of dwarf culture orchards harvest. A new hybrid visual servoing adaptive tracking controller and three adaptive laws are proposed to guarantee harvesting robots to finish the dynamic harvesting task and the adaption to unknown parameters including camera intrinsic and extrinsic model and robot dynamics. 
		By the Lyapunov theory, asymptotic convergence of the closed-loop system with the proposed control scheme is rigorously proven. Experimental and simulation results have been conducted to verify the performance of the proposed control scheme. The results demonstrate its effectiveness and superiority.
	\end{abstract}
	
	
	\section{Introduction}
	In fresh fruit production, harvest is one of the most labor-intensive operations and often depends on a large seasonal workforce, which is becoming less available. Harvesting robots are expected to cope with the non-customized and unstructured orchard working environment. 
	In the past years, many investigators have carried out manifold mechanisms and control strategies of robots to handle the complicated working environment\cite{ling2019dual,hohimer2019design,jiao2020detection}. Nowadays, an increasing number of orchards are employing modern agricultural strategies, high-density dwarf planting, also known as SNAP (Simple, Narrow, Accessible, and Productive) tree architectures, bringing considerable convenience to robots as well as humans in harvest operation\cite{bloch2018methodology,LEVIN2019104987}. 
	On this basis, in this paper, we focus on the investigation of the control strategy of harvesting robots in SNAP orchards, which is characterized by fast approaching, dynamic tracking, and robustness of model parameters. 
	
	One of the distinct characteristics of SNAP orchards is dense growth. It leads to a more significant interaction of adjacent fruits while harvesting. Fruits are prone to be in motion when the branches are pulled or dragged by harvesters, leading to difficulties for robots aligning objects. Consequently, the possibility of unsuccessful picks increases, and meanwhile the overall harvesting efficiency falls. In conclusion, the capability of robots to handle the dynamic motion of the target fruits is of significance. 
	
	For a long time, cameras have been employed in robotic systems to enhance their sensory ability. In the field of agricultural robots, many vision-based robotic control systems are open-loop structure\cite{bac2014harvesting,zhang2020state}. Despite the simplicity of the open-loop control, these systems may suffer from excessive positioning errors in outdoor agricultural environments since continuous image feedback is not opted to verify and rectify the position of the robot w.r.t. the fruit.
	
	As an alternative, a visual servoing scheme ensures a better performance with dynamic objects by combining real-time visual feedback and the actuator control \cite{chaumette2007visual}. It makes robots capable of updating objects' locations while an end-effector moves toward fruit in real-time\cite{li2019adaptive}. 
	\cite{mehta2014vision} presents a hybrid VS controller combining IBVS and PBVS to control the 3D translation of the camera to guarantee exponential regulation of the robot to a target fruit. 
	Due to environmental disturbances, the fruit target could be in motion. To address this problem, \cite{mehta2016adaptive} proposed an adaptive VS controller for harvesting robots to compensate for unknown dynamics of fruits online, and afterward, \cite{mehta2020finite} presents a finite-time VS controller to improve robustness of a robot system w.r.t. fruit motion. 
	
	Although many positive results improve the real-time performance of harvesting robots in practical orchards, there are still technical challenges in this field.
	
	The first one is the modeling of fruit motion. In the existing literature, the fruit motion is formulated by a combination of two second-order spring-mass systems, e.g. Eq.$(3)$ in \cite{mehta2016adaptive} and is regarded as known dynamics, e.g. Eq.$(11)$ in \cite{mehta2020finite}. Such a hypothesis makes sense in some situations but brings conservativeness that limits its performance in practical application.
	
	The second challenge is the convergence of visual servoing control systems. Many existing papers concentrate on improving the quality of stability with various methods, such as exponential stability\cite{mehta2014vision}, finite-time stability \cite{mehta2016adaptive}. The common control objective of the above work, however, is regulating the end-effector to the desired pixel coordinates and desired fruit depth, e.g.,  Eq.$(5)$ in \cite{mehta2020finite}. In the set-point control, the velocity and trajectory of the end-effector approaching the target cannot be regulated. In many cases, users may want the end-effector to perform a smooth and variable speed when approaching the fruit or to go along a specific path.
	
	The third challenge is parameter calibration. 
	A typical way to acquire camera parameters is usually by camera calibration which has been extensively discussed in literature \cite{heikkila2000geometric}. Moreover, both the calibration of systematic kinematic parameters and the robot dynamic parameters are tedious and inconvenient as well, especially for harvesting robotic systems.
	
	Basing on the three challenges mentioned above, in this paper, we investigate a hybrid visual servoing control scheme for harvesting robots in SNAP orchards. The contributions lie in the three aspects. a) We consider the unknown fruit motion by proposing a hybrid visual servoing configuration. In this paper, an eye-in-hand RGBD camera and a fixed RGBD camera are used to provide abundant image features. b) Visual servoing tracking control is investigated. The desired image trajectory is time-varying. It allows the users to define the way of end-effector approaching fruits.
	c) Non-calibration. With unknown camera extrinsic and intrinsic parameters and robotic dynamic parameters, the proposed VS tracking controller is uncalibrated. Three adaptive laws are also proposed to update unknown parameters online.

	\section{Model Formulation}\label{modelformulation}
	
	\subsection{Hybrid visual servoing system}
	
	For an RGBD camera, the following relationship holds
	
	\begin{equation}\label{Image-Cartersian-map-rgbd}
	{\mathbf{y}}_{rgbd}=
	\Omega_{rgbd}{\mathbf{x}^c}, 
	\end{equation}
	where $\mathbf{x}^c(t) \in \mathbb{R}^{3}$ denotes the feature point coordinates of the object in Cartesian frame $\mathcal{F}$ of the RGBD camera; $\mathbf{y}_{rgbd}(t)\in \mathbb{R}^{3}$ denotes the camera image coordinates on the image-space, which consists of two pixel positions on the RGB image and the depth value on the depth image; $\Omega_{rgbd}\in \mathbb{R}^{3\times3}$ denotes the intrinsic parameters matrix of RGBD cameras which satisfies
	\begin{equation}\label{Omega-rgbd}
	{\Omega}_{rgbd} = \begin{bmatrix}
	fk_u/z&fk_u\cot\vartheta/z&u_0/z\\
	0&\frac{fk_v}{z\sin\vartheta}&v_0/z\\
	0&0&\mu
	\end{bmatrix}.
	\end{equation}
	For the detail of matrix $\Omega_{rgbd}$, please refer to \cite{fuchs2008extrinsic,liu2006uncalibrated}. Here, $\mu$ denotes the approximation relationship between the depth measurement and its true value.

	In a hybrid VS system, two RGBD cameras are employed for eye-in-hand and eye-to-hand vision and their image coordinates vectors can be defined by $\mathbf{y}_{\text{eih}}$ and $\mathbf{y}_{\text{fixed}}$ respectively. Likewise, their instrinsic parametric matrices are denoted by ${\Omega}_{\text{eih}}$, ${\Omega}_{\text{fixed}}$.
	
	Once finished mapping feature positions from camera image-space to camera Cartesian-space, the relationships between manipulators and cameras are required. In hybrid VS systems, the following homogeneous transformation holds:
	
	\begin{equation}\label{eih-fixed-map}
	\sideset{^{\text{eih}}}{_{\text{fixed}}}{\mathop{\mathbf{T}}} = \sideset{^{\text{eih}}}{_{\text{EE}}}{\mathop{\mathbf{T}}} \sideset{^{\text{EE}}}{_{\text{BASE}}}{\mathop{\mathbf{T}}} 
	\sideset{^{\text{BASE}}}{_{\text{fixed}}}{\mathop{\mathbf{T}}},
	\end{equation}
	where $\sideset{^{\text{eih}}}{_{\text{EE}}}{\mathop{\mathbf{T}}}\in\text{SE(3)}$ denotes the homogeneous transformation matrix from the eye-in-hand camera frame to the end-effector frame; $\sideset{^{\text{EE}}}{_{\text{BASE}}}{\mathop{\mathbf{T}}} \in\text{SE(3)}$ denotes the matrix from the end-effector frame to the base frame; $\sideset{^{\text{BASE}}}{_{\text{fixed}}}{\mathop{\mathbf{T}}}\in\text{SE(3)} $ denotes the matrix from the base frame to the eye-to-hand camera frame.
	
	By the above transformation, the mapping relation of feature point coordinates $\mathbf{x}^c(t)$ between different camera frames in Cartesian-space can be formulated as follows 
	\begin{equation}\label{eih-fixed-carte-map}
	{\mathbf{x}}^{\text{eih}} = \sideset{^{\text{eih}}}{_{\text{fixed}}}{\mathop{\mathbf{T}}} {\mathbf{x}}^{\text{fixed}},
	\end{equation}
	where ${\mathbf{x}}^{\text{eih}},{\mathbf{x}}^{\text{fixed}}\in \mathbb{R}^4$ are vectors with 4 elements that the 4th element is $1$ to align the homogeneous transformation.
	Also the following mapping relation hold:
	\begin{equation}\label{eih-ee-map}
	{\mathbf{x}}^{\text{eih}} = \sideset{^{\text{eih}}}{_{\text{EE}}}{\mathop{\mathbf{T}}} {\mathbf{x}}^{\text{EE}}.
	\end{equation}
	Generally speaking, the relative pose between the eye-in-hand camera and the end-effector is invariant as well as the eye-to-hand camera and the robot base during the operation of the manipulator.

	Substituting Eq.\eqref{eih-ee-map} into Eq.\eqref{Image-Cartersian-map-rgbd}, one has 
	\begin{equation}\label{eih-fixed-image-carte-map}
	{\mathbf{y}}_{\text{eih}} = \Omega_{\text{eih}} \sideset{^{\text{eih}}}{_{\text{EE}}}{\mathop{\mathbf{T}}} {\mathbf{x}}^{\text{EE}} = \mathbf{M}_{\text{eih}}{\mathbf{x}}^{\text{EE}},
	\end{equation}
	where $\mathbf{M}_{\text{eih}} \in \mathbb{R}^{3\times4}$ denotes perspective projection matrix which is invariant.
	
	Differentiating Eq.\eqref{eih-fixed-image-carte-map} and combining them, one has
	\begin{equation}\label{eih-fixed}
	\begin{aligned}
	\dot{{\mathbf{y}}}_{\text{eih}} = 
	&\mathbf{M}_{\text{eih}} \sideset{^{\text{EE}}}{_{\text{BASE}}}{\mathop{\dot{\mathbf{T}}}} \mathbf{M}_{\text{fixed}}^+{\mathbf{y}}_{\text{fixed}}\\
	& + \mathbf{M}_{\text{eih}} \sideset{^{\text{EE}}}{_{\text{BASE}}}{\mathop{\mathbf{T}}} \mathbf{M}_{\text{fixed}}^+\dot{{\mathbf{y}}}_{\text{fixed}}
	\end{aligned}
	\end{equation} 
	where $\mathbf{M}_{\text{fixed}}^+$ is the pseudo-inverse matrix of $\mathbf{M}_{\text{fixed}}$.
	
	Rewriting Eq.\eqref{eih-fixed}, one has
	\begin{equation}\label{Jacobian-1}
	\begin{aligned}
	\dot{{\mathbf{y}}}_{\text{eih}} = 
	\mathbf{Q}\dot{\mathbf{q}}  + \mathbf{J}\dot{\mathbf{y}}_{\text{fixed}} 
	\end{aligned}
	\end{equation}
	where $\dot{\mathbf{q}}\in \mathbb{R}^n$ denotes the joint velocity vector;  $\mathbf{Q} \in \mathbb{R}^{3\times n}$ is robot Jacobian matrix, which describes the forwoard differential kinematics between image-space of eye-in-hand cameras and joint-space of the robot; $n$ denotes the dimensions of joint-space; $\mathbf{J}\in \mathbb{R}^{3\times 3}$ describes the differential kinematic relation between image-space of the eye-in-hand camera and the fixed camera.
	
	\begin{property}\label{pp1}
		For a vector $\phi\in \mathbb{R}^{n\times 1}$, the products $\mathbf{Q}\phi$ can be linearly parameterized as follows
		\begin{equation}\label{Prop1-eq}
		\mathbf{Q}\phi=\mathbf{Y}(\mathbf{y}_{\text{fixed}},\mathbf{q},\phi)\theta_k,
		\end{equation}
		where $\mathbf{Y}(\mathbf{y}_{\text{fixed}},q,\phi) \in \mathbb{R}^{3\times p_1}$ are regressor matrices which consist of the known parameters; $\theta_k \in \mathbb{R}^{p_1\times 1}$ is a vector which consists of unknown parameters. The number of unknown parameters $p_1$ varies according to the D-H parameters of the robot. 
	\end{property}
	
	\begin{property}\label{pp2}
		For a vector $\phi\in \mathbb{R}^{3\times 1}$, the products $\mathbf{J}\phi$ can be linearly parameterized as follows
		\begin{equation}\label{Prop2-eq}
		\mathbf{J}\phi=\mathbf{W}(\mathbf{q},\phi)\theta_m,
		\end{equation}
		where $\mathbf{W}(q,\phi) \in \mathbb{R}^{3\times p_2}$ are regressor matrices which consist of the known parameters; $\theta_h \in \mathbb{R}^{p_2\times 1}$ is a vector which consists of unknown parameters. The number of unknown parameters $p_2$ varies according to the D-H parameters of the robot. 
	\end{property}

	\subsection{Robotic dynamics}
	It is well-known that the dynamics of a robot manipulator can be formulated as \cite{Khalil2002Nonlinear}
	\begin{equation}\label{10}
	\begin{aligned}
	\mathbf{H}(\mathbf{q}(t))\ddot{\mathbf{q}}(t)+\left(\frac{1}{2}\dot{\mathbf{H}}(\mathbf{q}(t))+\mathbf{C}({\mathbf{q}}(t),\dot{\mathbf{q}}(t))\right)\dot{\mathbf{q}}+&\mathbf{g}(\mathbf{q}(t))\\
	&=\tau,
	\end{aligned}
	\end{equation}
	where $\tau$ is the $n\times 1$ joint input of the manipulator, $\mathbf{H}({\mathbf{q}}(t))$ is the $n\times n$ positive-define and symmetric inertia matrix and $\mathbf{C}({\mathbf{q}}(t),\dot{\mathbf{q}}(t)) \in \mathbb{R}^{n\times n}$ is a skew-symmetric matrix such that for any proper dimensional vector $\psi$, 
	\begin{equation}\label{pp-dyn}
	\psi^T\mathbf{C}({\mathbf{q}}(t),\dot{\mathbf{q}}(t))\psi=0.
	\end{equation}
	On the left side of (\ref{10}), the first term is inertia force, the second term represents the Coriolis and centrifugal forces, and the last term $\mathbf{g}(\mathbf{q})$ represents the gravitational force.
	Eq.\eqref{10} can be expressed in a linearizing parameterized form and satisfies the following property \cite{su2009global}.
	\begin{property}\label{property4}
		The dynamic equation of robot manipulator can be expressed as a linear function as follow
		\begin{equation}\label{Pdy1}
		\begin{aligned}
		\mathbf{H}(\mathbf{q}(t))\ddot{\mathbf{q}}(t)+\left(\frac{1}{2}\dot{\mathbf{H}}(\mathbf{q}(t))+\mathbf{C}({\mathbf{q}}(t),\dot{\mathbf{q}}(t))\right)\dot{\mathbf{q}}+\mathbf{g}(\mathbf{q}(t))&\\
		=\mathbf{Y}_d(\mathbf{q},\dot{\mathbf{q}},\dot{\mathbf{q}}(t),\ddot{\mathbf{q}}(t))\theta_d&,
		\end{aligned}
		\end{equation}
		where $\xi \in \mathbb{R}^{n\times 1}$, $\mathbf{Y}_d(\mathbf{q},\dot{\mathbf{q}},\dot{\mathbf{q}}(t),\ddot{\mathbf{q}}(t)) \in \mathbb{R}^{n\times p_3}$ is the corresponding dynamic regressor matrix and $\theta_d \in \mathbb{R}^{p_3 \times 1}$ denotes the unknown dynamic parameter vector. The number of unknown parameters is denoted as $p_3$ and the value of $p_3$ depends on the number of the joint dimensions \cite{chang2020toward}.
	\end{property}

	\section{Hybrid dynamic visual tracking control and stability analysis}\label{sec controller}
	
	\subsection{Problem description}\label{control-obj}
	In the system of this paper, three prerequisites are considered, summarized as follows.
	\begin{enumerate}
		\item[a)]  The target fruit is in motion w.r.t. the robot base.
		\item[b)]  The camera intrinsic and extrinsic parameters and robot dynamic parameters are unknown.
		\item[c)]  The desired image state is a dynamic trajectory rather than a static position.
	\end{enumerate}
	The control objective is to guarantee the end-effector of the robot to track a user-defined desired trajectory approaching a target fruit for finishing harvesting, which can be formulated by
	\begin{equation}\label{objective}
	\lim_{t\rightarrow\infty}\Delta\mathbf{y},\Delta\dot{{\mathbf{y}}}=0,	
	\end{equation}
	where $	\Delta{{\mathbf{y}}} = {{\mathbf{y}}} - {{\mathbf{y}}}_d$,  $	\Delta{\dot{\mathbf{y}}} = {\dot{\mathbf{y}}} - {\dot{\mathbf{y}}}_d$ and $({{\mathbf{y}}}_d,\dot{{\mathbf{y}}}_d,\ddot{{\mathbf{y}}}_d)$ is the desired image trajectory that is defined in the eye-in-hand RGBD camera. Note that, for the purpose of simplification, hereafter, we remove the subscript "eih" and use ${{\mathbf{y}}}$ to be the shorthand of ${{\mathbf{y}}}_{\text{eih}}$
	but keep the subscript "fixed" in ${{\mathbf{y}}}_{\text{fixed}}$ to tell apart. 
	\subsection{Controller design}
	We first introduce the following reference image velocity
	\begin{equation}\label{ref-image-velocity}
	\dot{{\mathbf{y}}}_r = \dot{{\mathbf{y}}}_d - \lambda({{\mathbf{y}}} - {{\mathbf{y}}}_d)-\hat{\mathbf{J}}\dot{{\mathbf{y}}}_{\text{fixed}},
	\end{equation}
	where $\lambda$ is a positive constant; $\hat{\mathbf{J}}$ denotes the estimation matrix of  ${\mathbf{J}}$ that satisfies
	\begin{equation}\label{jhat}
	\hat{\mathbf{J}}\phi=\mathbf{W}(\mathbf{q},\phi)\hat{\theta}_m,
	\end{equation}
	where $\hat{\theta}_m$ is the estimation of unknown vector ${\theta}_m$.
	By \eqref{ref-image-velocity}, a novel joint-space reference velocity is defined by
	\begin{equation}\label{qr}
	\dot{\mathbf{q}_r} =\hat{\mathbf{Q}}^+\dot{{\mathbf{y}}}_r,
	\end{equation}
	where $\hat{\mathbf{Q}}^+$ denotes the pseudo-inverse of estimated Jacobian matrix and $\hat{\mathbf{Q}}$ is derived from
	\begin{equation}\label{qhat}
	\hat{\mathbf{Q}}\phi=\mathbf{Y}(\mathbf{y}_{\text{fixed}},\mathbf{q},\phi)\hat{\theta}_k,
	\end{equation}
	where $\hat{\theta}_k$ is the estimation of unknown vector ${\theta}_k$. Note that both $\hat{\theta}_k$ and $\hat{\theta}_m$ are online updated by the adaptive laws that will be proposed later.
	
	Then a sliding vector in joint-space can be constructed as follows:
	\begin{equation}\label{sq}
	\mathbf{s}_q = \dot{\mathbf{q}} - \dot{\mathbf{q}_r},
	\end{equation}
	
	Now we are in a position to present the tracking VS controller of this paper
	
	\begin{equation}\label{controller}
	\tau = \mathbf{Y}_d(\mathbf{q},\dot{\mathbf{q}},\dot{\mathbf{q}}_r,\ddot{\mathbf{q}}_r)\hat{\theta}_d-\hat{\mathbf{Q}}^T\mathbf{K}_1\Delta\mathbf{y} - \mathbf{K}_2\mathbf{s}_{\mathbf{q}},
	\end{equation}
	where $\mathbf{K}_1, \mathbf{K}_2 \in \mathbb{R}^{n\times n}$ are positive definitely symmetric matrices to be determined and $\hat{\theta}_d \in \mathbb{R}^{p_3 \times 1}$ denotes the estimation of the unknown dynamic parameter vector.

	By combining Eq.\eqref{Prop1-eq} and Eq.\eqref{qhat}, the estimation error of robot Jacobian matrix is
	\begin{equation}\label{jac-error}
	\hat{\mathbf{Q}}\dot{\mathbf{q}} - {\mathbf{Q}}\dot{\mathbf{q}} = \mathbf{Y}(\mathbf{y}_{\text{fixed}},\mathbf{q},\dot{\mathbf{q}})\Delta\hat{\theta}_k.
	\end{equation}
	Likewise, with Eq.\eqref{Prop2-eq} and Eq.\eqref{jhat}, one has 
	\begin{equation}\label{vis-map-error}
	\hat{\mathbf{J}}\dot{\mathbf{y}}_{\text{fixed}} -{\mathbf{J}}\dot{\mathbf{y}}_{\text{fixed}}  = \mathbf{W}(\mathbf{q},\dot{\mathbf{y}}_{\text{fixed}})\Delta\theta_m.
	\end{equation}
	By combining Eq.\eqref{jac-error}, Eq.\eqref{qr} and Eq.\eqref{ref-image-velocity}, one has
	\begin{equation}\label{Qsq}
	\hat{\mathbf{Q}}\mathbf{s}_{\mathbf{q}}=\Delta \dot{\mathbf{y}} + \lambda \Delta\mathbf{y} + \Delta{\mathbf{J}}_v\dot{\mathbf{y}}_{\text{fixed}}+\mathbf{Y}_k(\mathbf{y}_{\text{fixed}},\mathbf{q},\dot{\mathbf{q}},\dot{\mathbf{q}}_r)\Delta\theta_k.
	\end{equation}
	
	According to Eq.\eqref{Pdy1}, it yields
	\begin{equation}\label{dynamic-sq}
	\begin{aligned}
	&\mathbf{Y}_d(\mathbf{q},\dot{\mathbf{q}},\dot{\mathbf{q}}(t),\ddot{\mathbf{q}}(t))\theta_d - \mathbf{Y}_d(\mathbf{q},\dot{\mathbf{q}},\dot{\mathbf{q}}_r,\ddot{\mathbf{q}}_r)\hat{\theta}_d\\
	=&\mathbf{H}(\mathbf{q}(t))\dot{\mathbf{s}}_q+\left(\frac{1}{2}\dot{\mathbf{H}}(\mathbf{q}(t))+\mathbf{C}({\mathbf{q}}(t),\dot{\mathbf{q}}(t))\right){\mathbf{s}}_q\\
	&-\mathbf{Y}_d(\mathbf{q},\dot{\mathbf{q}},\dot{\mathbf{q}}_r,\ddot{\mathbf{q}}_r)\Delta{\theta}_d.
	\end{aligned}	
	\end{equation}
	
	\subsection{Adaptive laws}
	In the above analysis, the estimation of $\hat{\mathbf{Q}},\hat{\mathbf{J}}$ and robotic dynamics \eqref{Pdy1} are used. These estimations can be linearized by Property \ref{pp1}-\ref{property4} and can be obtained by the estimations of vectors, $\hat{\theta}_k,\hat{\theta}_m,\hat{\theta}_d$. In order to obtain these estimates of the unknown vectors, we present the following adaptive laws:
	\begin{equation}\label{dynamic-laws}
	\dot{\hat{\theta}}_d = -\Psi_d^{-1}\mathbf{Y}^T_d(\mathbf{q},\dot{\mathbf{q}},\dot{\mathbf{q}}_r,\ddot{\mathbf{q}}_r)\mathbf{s}_q,
	\end{equation}
	\begin{equation}\label{kinematic-laws}
	\dot{\hat{\theta}}_k = 
	\Psi_k^{-1}\mathbf{Y}^T_k(\mathbf{y}_{\text{fixed}},\mathbf{q},\dot{\mathbf{q}},\dot{\mathbf{q}}_r)\mathbf{K}_1\Delta\mathbf{y},
	\end{equation}
	\begin{equation}\label{kinematic-vis-laws}
	\dot{\hat{\theta}}_m = 
	\Psi_m^{-1}\mathbf{W}^T(\mathbf{q},\dot{\mathbf{y}}_{\text{fixed}})\mathbf{K}_1\Delta\mathbf{y},
	\end{equation}
	where $\Psi_d,\Psi_k$ and $\Psi_m$ are positive definite symmetric matrices with proper dimensions.

	\section{Experimental and Simulating Results}\label{sec_exp}
	The experimental platform of apple perception consists of two Realsense D435i and NVIDIA Jetson TX2. The image processing system is based on the algorithm of YOLOv5\cite{Yolo5} to real-time detect fruits in RGB frames. 
	
\begin{figure}[h]
		\centering
		\includegraphics[width=\linewidth]{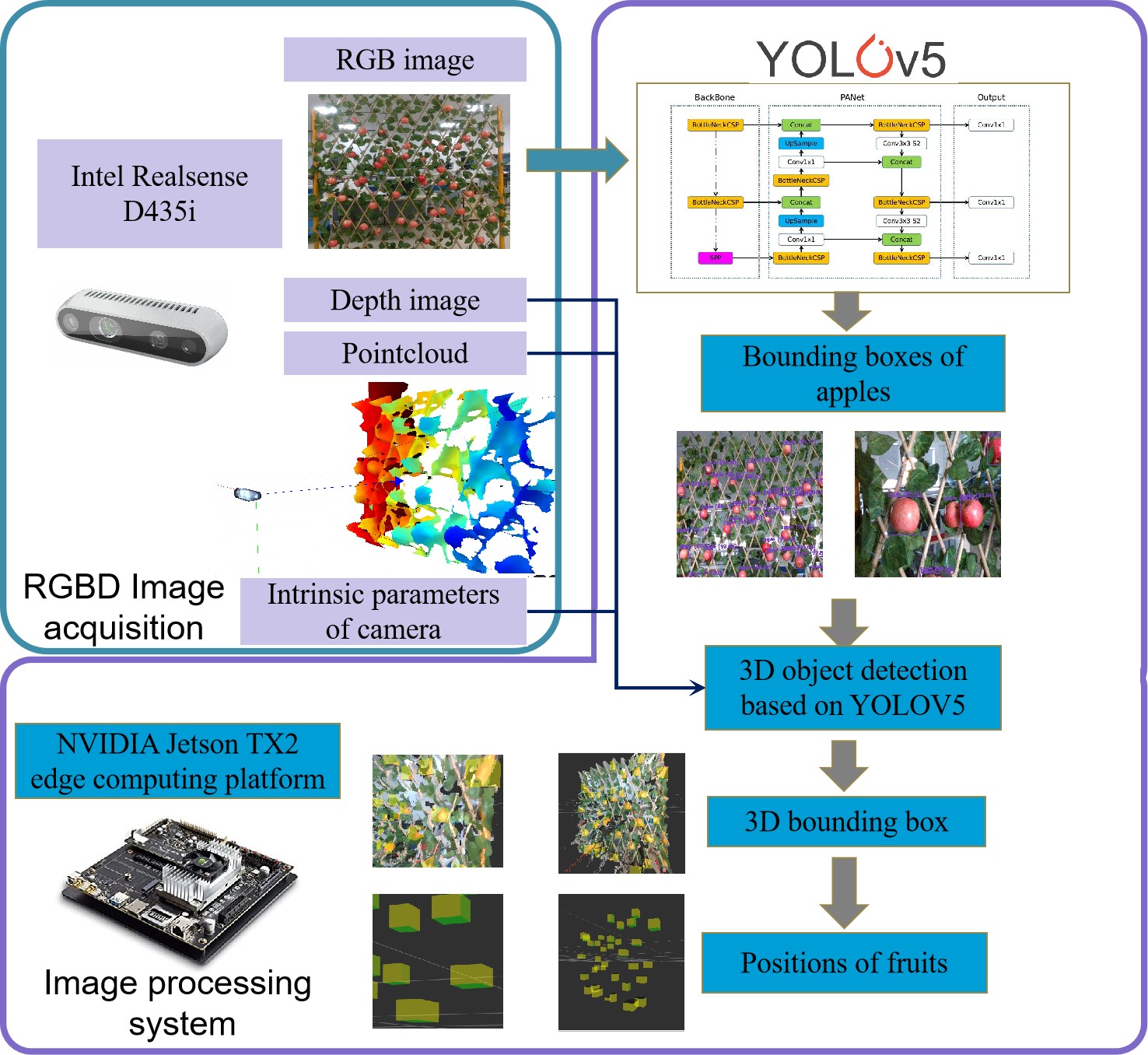}
		\caption{3D real-time apple detection system based on YOLOv5}
		\label{fig:3dyolov5}
	\end{figure}

\begin{figure}[h!]
		\centering
		\subfloat[Detection results of the fixed RGBD camera]{\includegraphics[width = 0.7\linewidth]{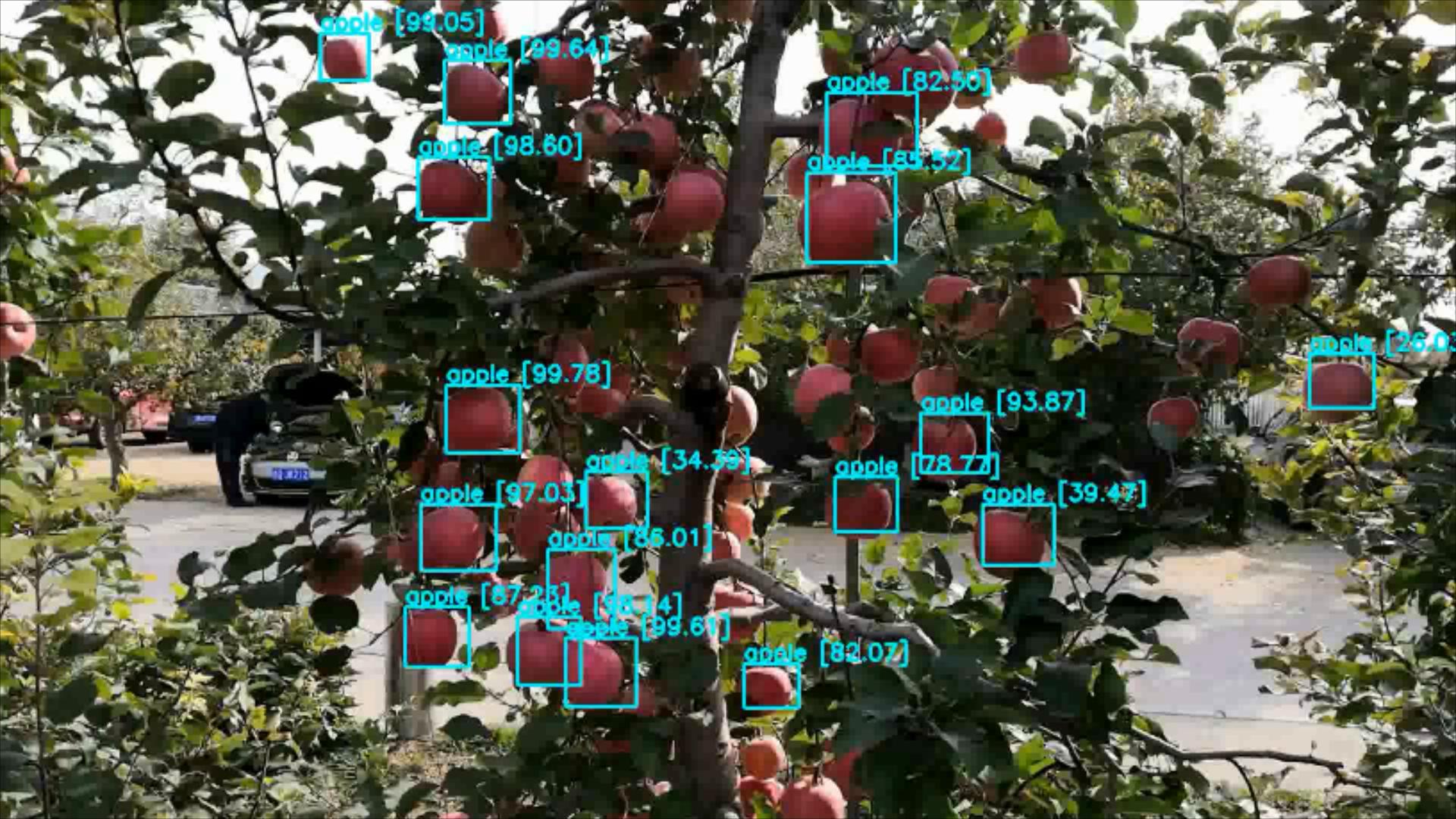}}
		
		\centering
		\subfloat[Detection results of the eye-in-hand RGBD camera]{\includegraphics[width = 0.7
			\linewidth]{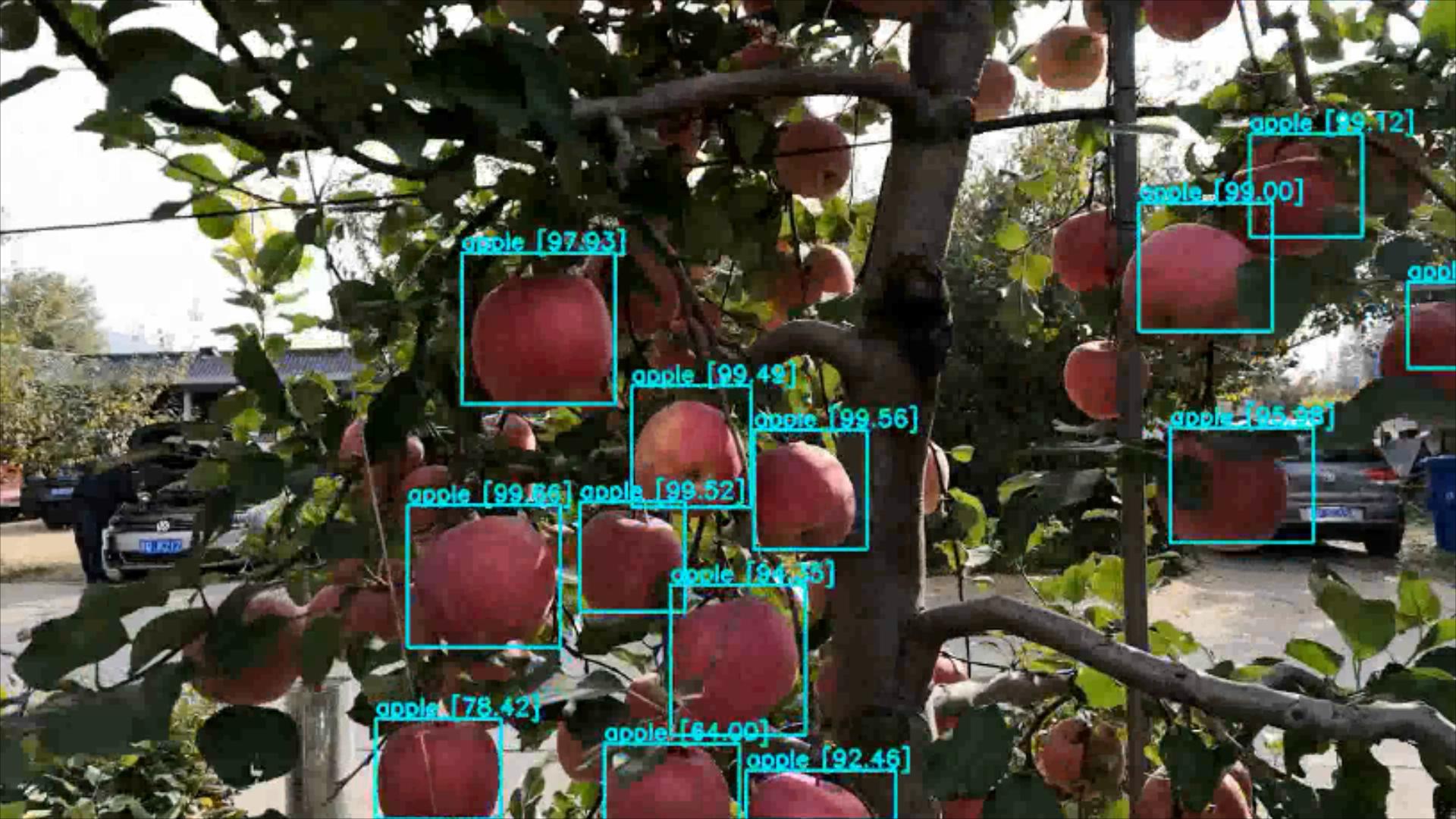}}
		\caption{Fruits detection experimental results}
		\label{fig:orcharddetection}
	\end{figure}

	Fig.\ref{fig:3dyolov5} demonstrates the system of 3D real-time apple detection based on YOLOv5. There are two RGBD cameras mounted on the end-effector and a fixed position of the robot base respectively. Once finished RGBD image acquisition, the image computing system, i.e. edge computing modules, processes RGB images and then depth images based on the YOLOv5 network and calculates the depth parameters from 2D bounding boxes. Combining with 2D bounding boxes, depth parameters and intrinsic parameters of cameras, the final outputs of image processing systems are 3D Cartesian positions of fruits in RGBD cameras. In the experiments, the quantitative performance measures of the proposed YOLOv5-based 3D detection algorithm are as follows: mAP: 0.91, recall: 0.88, accuracy: 0.913, IoU: 0.873, speed: 23FPS.
	The experimental results in orchards are shown in Fig.\ref{fig:orcharddetection}. It can be seen that apple targets are detected in the different conditions (light, shelter, et al.). 
	
	We use a quasi digital twin method to establish a digital replica of a dwarf culture orchard. To this end, MuJoCo (Multi-Joint dynamics with Contact), a physics engine, is employed to facilitate the simulation.

\begin{figure}[h]
		\centering
		\includegraphics[width=\linewidth]{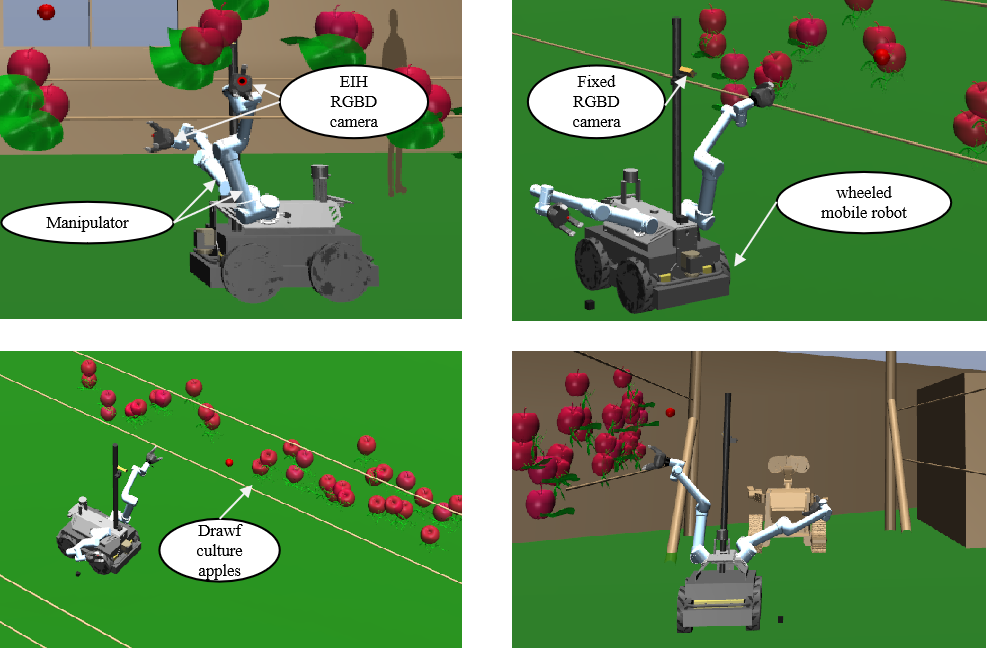}
		\caption{The simulating harvesting environment based on MuJoCo}
		\label{fig:mujoco-env}
	\end{figure}

	In this simulation, we model an orchard working environment with two-arm harvesting robots, as shown in Fig.\ref{fig:mujoco-env}. The fixed RGBD camera is installed at a metal stand which is immobilized to the wheeled mobile robot and the EIH RGBD camera is mounted at the end-effector. The apples are modeled by apple-shaped mesh files and are flexibly connected to different junctions simulating apples hanging on branches. Besides, the apples in the virtual environment are also assigned physical properties, such as gravity, friction, collision et al. The two arms used are UR5 and are modeled by the official urdf files.

	\begin{figure}[h]
		\centering
		\subfloat[The proposed controller]{\includegraphics[width = 0.7\linewidth]{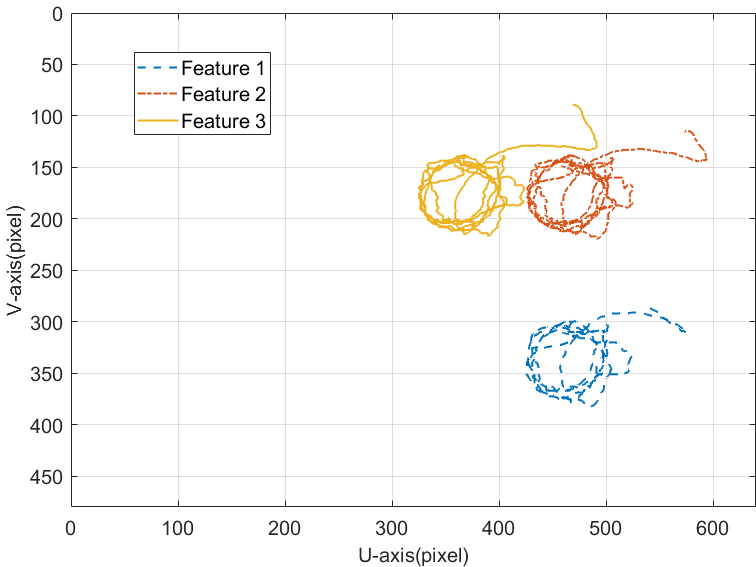}\label{fig:result-2-round-a}}
		
		\centering
		\subfloat[The reference controller]{\includegraphics[width = 0.7
			\linewidth]{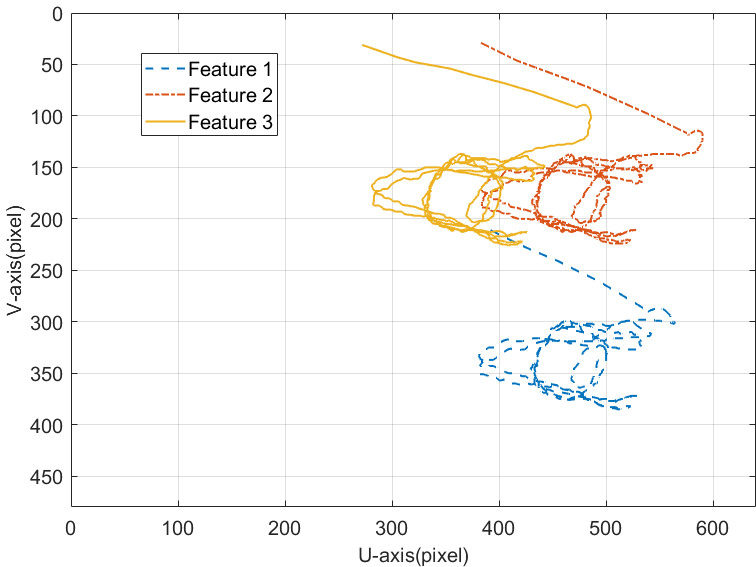}\label{fig:result-2-round-b}}
		\caption{Actual image trajectories of feature points under two control schemes with circular motions of a target}
		\label{fig:result-2-round}
	\end{figure}
	
	\begin{figure}[h]
		\centering
		\subfloat[The RGB image errors convergence of Feature 1]{\includegraphics[width = 0.7\linewidth]{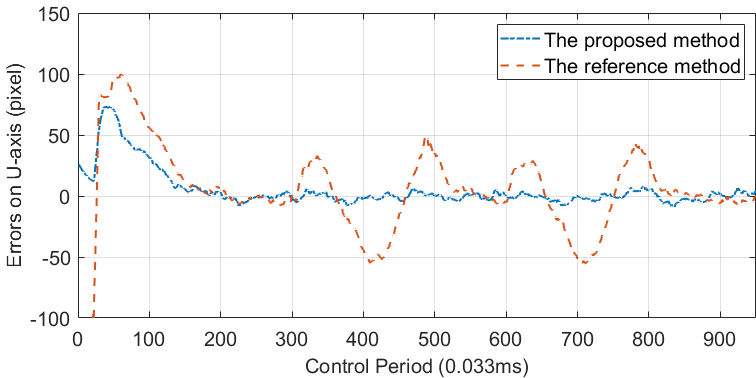}\label{fig:result-3-round-a}}
		
		\centering
		\subfloat[The torque outputs of the proposed controller]{\includegraphics[width = 0.7
			\linewidth]{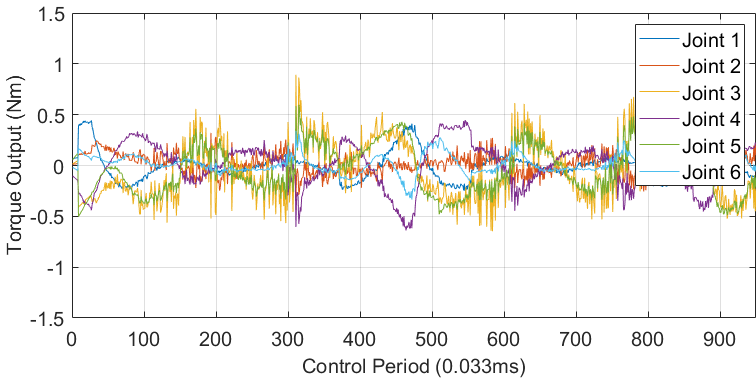}\label{fig:result-3-round-b}}
		\caption{The profiles of image errors and the controller torque outputs of the proposed control method in the experiments of cicular motion}
		\label{fig:result-3-round}
	\end{figure}
	
	\begin{figure}[h!]
		\centering
		\subfloat[The proposed controller]{\includegraphics[width = 0.7\linewidth]{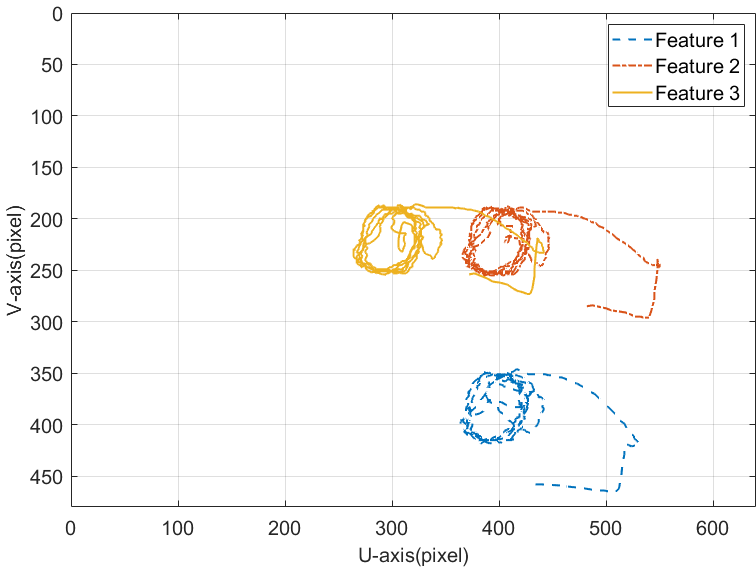}\label{fig:result-2-fang-a}}
		
		\centering
		\subfloat[The reference controller]{\includegraphics[width = 0.7
			\linewidth]{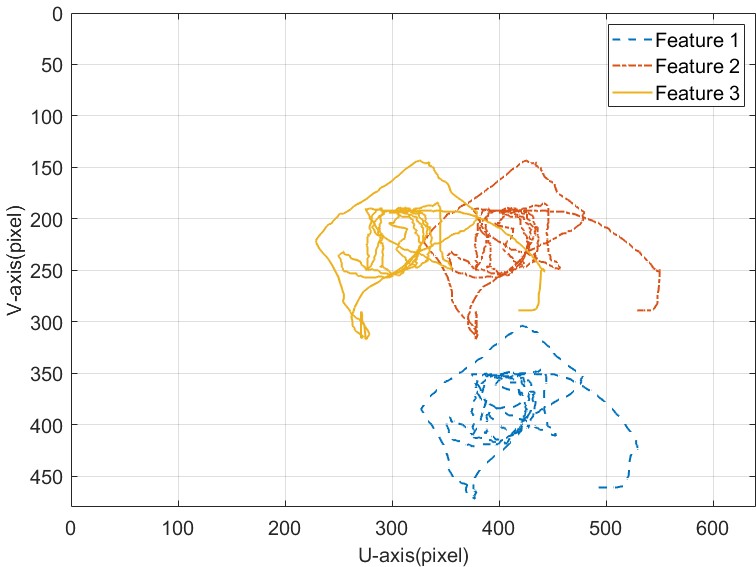}\label{fig:result-2-fang-b}}
		\caption{Actual image trajectories of feature points under two control schemes with rectangle motions of a target}
		\label{fig:result-2-fang}
	\end{figure}
	
	\begin{figure}[h!]
		\centering
		\subfloat[The RGB image errors convergence of Feature 1]{\includegraphics[width = 0.7\linewidth]{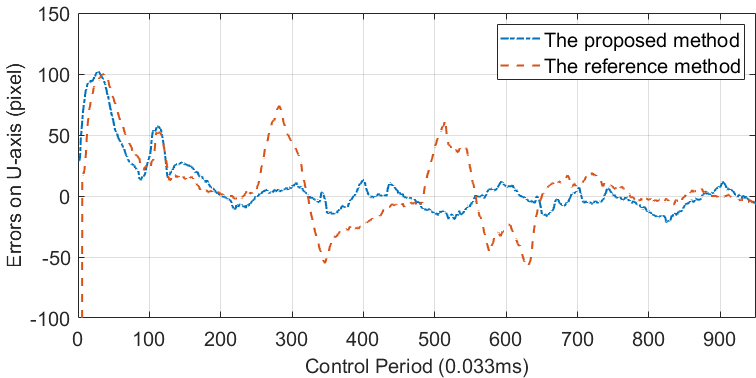}\label{fig:result-3-fang-a}}
		
		\centering
		\subfloat[The torque outputs]{\includegraphics[width = 0.7
			\linewidth]{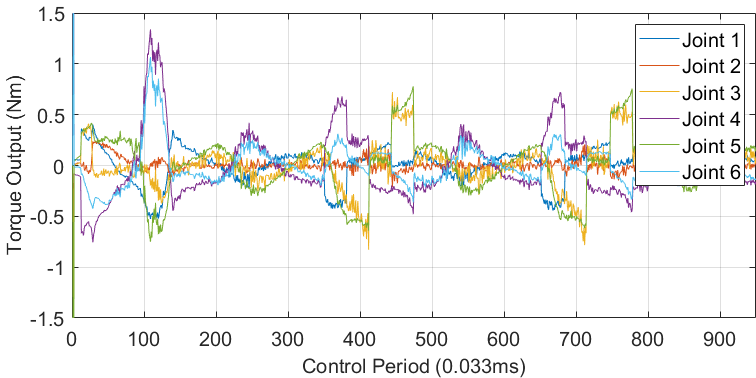}\label{fig:result-3-fang-b}}
		\caption{The profiles of image errors and the controller torque outputs of the proposed control method in the experiments of rectangle motion}
		\label{fig:result-3-fang}
	\end{figure}

	Recalling the problem description in subsection \ref{control-obj}, we clarify how the three prerequisites are defined in the simulation. The motions of a target fruit w.r.t. the robot base are considered in two ways: circle and rectangle. The circular motion resembles the apple swinging on the branch; the rectangle motion resembles that the harvesting robot is performing search behavior along the fruit wall, where the rectangle motion is caused by the motion of the robot base. Second, the camera intrinsic and extrinsic parameters and robot dynamic parameters are not used in the controller design even though they can be easily obtained in our simulation. Yet they are not available all the time in practice. Third, the desired image trajectory is defined by a cylindrical spiral in the 3-dimensional image space of the RGBD camera. Note that in practical harvest, the approaching way along with a cylindrical spiral maybe not an efficient one. We use it just for demonstrating the tracking performance better.

	Based on the above analysis, the simulations are conducted to verify the tracking performance of the proposed control method with the different motion of target features and with uncalibrated model parameters. 
	To verify the superiority of the proposed hybrid VS configuration, a controller proposed by \cite{li2018visual} with EIH configuration is taken as the reference.
	Moreover, for the visual servoing tasks, at least three points are necessary to construct a 6-dimensional Jacobian matrix \cite{4015997} to control 6DOF manipulators. Hence, we use three feature points of an apple to construct the visual servoing features. 
	
	The circular motion is first considered. In the settings of the controller \eqref{controller} and Jacobian matrices, we randomly assign initial values of the unknown parameters to verify the adaption of the proposed method. 
	Fig.\ref{fig:result-2-round-a} shows that the three feature points of a target can be tracked under the control of the proposed method; in contrast, Fig.\ref{fig:result-2-round-b} shows unsatisfactory trajectories by the reference control scheme.
	Fig.\ref{fig:result-3-round-a} presents that the image errors of Feature 1 with both the proposed method and the reference method. 
	Fig.\ref{fig:result-3-round-b} demonstrates the actual torque outputs of the proposed controller, demonstrating that the joint actuators can rapidly respond to the target's movement to guarantee the tracking convergence. For a rectangle motion of the robot base w.r.t. the target, the simulating results are given in Fig.\ref{fig:result-2-fang} and Fig.\ref{fig:result-3-fang}. 
	
	The results demonstrate that the proposed controller is capable to regulate the robots approaching targets along with the desired dynamic trajectory even with the circular and rectangle motions.

	\section{Conclusions}\label{sec con}
	
	In this paper, we proposed a new hybrid uncalibrated VS control scheme for harvesting robots to track the desired trajectory with the unknown moving feature point w.r.t. the base frame. A rigorous mathematic analysis is given to prove that the proposed controller guarantees the asymptotic convergence of the closed-loop system during the desired trajectory tracking in the presence of the threefold uncalibrated parameters. The performance of the proposed method has been verified by the comparative simulations.

\end{document}